\documentclass[a4paper,fleqn]{cas-dc}

\usepackage[numbers]{natbib}
\usepackage{multirow}

\def\tsc#1{\csdef{#1}{\textsc{\lowercase{#1}}\xspace}}
\tsc{WGM}
\tsc{QE}
\tsc{EP}
\tsc{PMS}
\tsc{BEC}
\tsc{DE}

\begin{document}
\let\WriteBookmarks\relax
\def\floatpagepagefraction{1}
\def\textpagefraction{.001}
\shorttitle{Instance-Aware KD for SSL of an On-Board MTL Model for Collision Avoidance System}
\shortauthors{G, Hwang et~al.}

\title [mode = title]{Instance-Aware Knowledge Distillation for Semi-Supervised Learning of an On-Board Multi-Task Dense Prediction Model for Collision Avoidance System}

\author[1]{Gyutae Hwang}[orcid=0000-0001-6365-2231]
\ead{gyutae741@jbnu.ac.kr}
\credit{Writing – review \& editing, Writing – original draft, Validation, Software, Methodology, Data curation, Conceptualization}

\author[1]{Sang Jun Lee\corref{cor1}}[orcid=0000-0002-9312-6299]
\ead{sj.lee@jbnu.ac.kr}
\credit{Writing – review \& editing}

\affiliation[1]{organization={Division of Electronics and Information Engineering, Jeonbuk National University},
            addressline={567 Baekje-daero, Deokjin-gu}, 
            city={Jeonju},
            postcode={54896},
            country={Republic of Korea}}

\cortext[cor1]{Corresponding author}

\begin{abstract}
Collision avoidance systems have evolved toward camera-based deep learning approaches for driving scene understanding.
However, deployment in edge environments such as country clubs is constrained by limited computational resources and unreliable communication infrastructure.
Moreover, constructing large-scale datasets for the target domain involves substantial annotation cost.
To address these limitations, we propose an instance-aware knowledge distillation framework for semi-supervised learning.
Specifically, we generate pseudo labels that mitigate teacher bias by leveraging domain priors from the teacher and instance-centric knowledge from foundation models.
The trained lightweight student is deployed in the proposed collision avoidance system and performs multiple dense prediction tasks in real-time.
The system detects frontal obstacles and encodes their spatial information into controller area network messages for automated guided vehicle operation.
To achieve this, we construct a large-scale country club dataset and perform field validation of the proposed system.
Experimental results demonstrate that the student outperforms the large teacher in instance segmentation while mitigating performance degradation in monocular depth estimation.
Compared with the teacher, the student reduces FLOPs by 22.68$\times$ and parameters by 14.33$\times$, achieving 6.46 FPS on a low-cost edge device.
\end{abstract}




\maketitle

\section{Introduction}
Collision avoidance systems are a primary component for the safety of most autonomous driving systems, aiming to recognize potential hazards through perception sensors.
Cameras represent the scene in a two-dimensional image space, providing comprehensive visual information of obstacles such as the size, perspective, and texture.
Recently, image-based deep learning techniques have demonstrated strong visual representation capability across various downstream tasks by leveraging large-scale general scene datasets and pretrained parameters \cite{zhang2024vision, uelwer2025survey}.
Deep learning-based collision avoidance systems employ downstream tasks such as monocular depth estimation, semantic segmentation, and object detection to recognize obstacles \cite{liu2023deep}.
However, computing resources in edge platforms are often constrained by limited power and physical size, which limits real-time model inference \cite{vu2023toward}.
Moreover, incorporating various downstream tasks increases the computational burden.
Another challenge is that edge platforms in specific domains require supervised fine-tuning of deep learning models, which incurs substantial labeling costs \cite{lee2025customkd}.

Country clubs (CC) are a representative outdoor domain where transportation platforms operate in edge environments, creating a practical need for collision avoidance systems.
CCs in Korea and Japan are generally located in mountainous areas and are characterized by wide sites and steep slopes.
Accordingly, automated guided vehicles (AGV) are used for transportation between holes.
Conventional AGVs are capable of unmanned operation based on ultrasonic and magnetic sensors, and they operate as line tracers that follow guide lines embedded in the pavement \cite{lee2024consideration}.
However, such approaches require high initial construction and maintenance costs, and even a failure in a short section can cause inconvenience to all users.
Moreover, CCs are often located far from telecommunications infrastructure and surrounded by dense vegetation, reducing the reliability of cloud-based services \cite{vougioukas2013influence, wang2024assessing}.
To address these limitations, we propose a camera-based collision avoidance system built on a low-cost edge device, along with a training framework for a lightweight deep learning model.

To build a camera-based collision avoidance system for edge environments, deep learning task integration and model compression are required.
Accurate risk assessment of obstacles in images demands pixel-level and instance-aware distance information derived from instance segmentation \cite{he2017mask, wang2020solo, bolya2019yolact} and monocular depth estimation \cite{eigen2014depth, lee2019big, bhat2021adabins}.
However, employing individual models for these two dense prediction tasks is computationally expensive and hinders the real-time operation of the system \cite{xu2023multi}.
Therefore, we designed an integrated deep learning model that performs both tasks in parallel utilizing multi-task learning (MTL).
MTL models employ a shared encoder and multi-decoder architecture designed to learn comprehensive task-agnostic representation \cite{crawshaw2020multi}.
However, achieving sufficient representation capability requires increasing the scale and complexity of the encoder network, which poses challenges for deployment in edge environments.

Besides model complexity, the high annotation cost of supervised learning poses another major challenge.
For depth estimation, ground truth (GT) depth maps can be generated automatically through the calibration of a camera and light detection and ranging (LiDAR).
In contrast, segmentation requires pixel-level annotation of irregular objects, resulting in a labor-intensive and time-consuming labeling process.
Semi-supervised learning (SSL) is a practical training approach that uses both labeled and unlabeled datasets to address limited GT data while reducing annotation costs, improving representation capability, and generalization performance \cite{lee2013pseudo, yang2022survey}.
Common SSL approaches include consistency regularization, entropy minimization, and self-training, among which pseudo labeling is the most widely used.
Pseudo labels are usually generated from the predictions of a prototype model trained on labeled dataset, and recent studies have focused on improving their quality \cite{tarvainen2017mean, sohn2020fixmatch}.

In this study, we propose a knowledge distillation (KD)-based training approach to address the computational and annotation costs of the presented real-world problem.
KD leverages the predictions of a large teacher model to train an efficient student model, reducing model complexity while preserving performance \cite{hinton2015distilling}.
The rich predictions of a large teacher model are well suited for pseudo label generation in SSL for a deployable student model. 
However, a teacher trained on a small labeled dataset often shows limited generalization ability to the target domain, resulting in object omission or incomplete pseudo masks on unlabeled data.
To address this limitation, we utilize foundation model outputs as auxiliary supervision sources that complement teacher predictions.
Recent dense prediction foundation models, such as segment anything model (SAM) \cite{kirillov2023segment}, and depth anything v2 (DAv2) \cite{yang2024depthv2}, have shown strong generalization ability across diverse visual patterns through large-scale pretraining.
We adopt a multi-teacher KD approach that utilizes the instance-centric knowledge of the large teacher and foundation models to generate CC domain pseudo labels and train a lightweight student model.

This study focuses on a real-time collision avoidance system for edge-based CC driving environments and proposes an instance-aware KD method for SSL of a lightweight MTL model.
As a result, the deployable MTL model achieved improved segmentation performance than a teacher model with over 14$\times$ higher complexity and outperformed the baseline SSL method in depth estimation.
The main contributions of this paper are as follows.
\begin{itemize}
    \item We collected a large-scale driving dataset specialized for country club environments
    \item We developed a real-time collision avoidance system based on a lightweight multi-task learning model
    \item We propose an instance-aware knowledge distillation method for semi-supervised learning to reduce both computing and annotation costs
\end{itemize}
The paper is organized as follows.
Section \ref{Related work} reviews related work.
Section \ref{Materials} introduces the materials, including the real-world dataset acquisition process and experimental equipment.
Section \ref{Methodology} presents the methodology, including the proposed system and semi-supervised learning pipeline.
Sections \ref{Experimental results} and \ref{Conclusion} present the experimental results and conclusion, respectively.

\section{Related work}
\label{Related work}
\subsection{Camera-based collision avoidance systems}
Recent deep learning-based collision avoidance systems have evolved toward recognizing objects from front-view images and assessing collision risk \cite{chitraranjan2025vision}.
Previous studies have primarily utilized object detection or semantic segmentation to identify the spatial location and semantic categories of objects \cite{awan2025deep, feng2020deep}.
Meanwhile, distance estimation methods have relied on explicit geometry-based approaches such as stereo camera disparity and monocular geometric estimation \cite{zaarane2020distance, ali2020real}.
Beyond perception algorithms, some studies have explored cloud-assisted deployment, where network communication with remote servers is used to obtain the results of large models in real-time \cite{vougioukas2013influence, groshev2023edge, gong2023edge}.
However, such approaches can be difficult to apply in CC environments where communication is often unreliable.
Conventional detection-based methods also have limitations in CC driving environments characterized by dense vegetation, steep slopes, irregular ground-surface objects, and narrow paths.
Moreover, rule-based distance estimation struggles to accurately measure distant objects in outdoor environments.
In contrast, this study aims to develop a low-cost collision avoidance system that relies only on a monocular camera and performs instance segmentation and depth estimation on an edge device.

\subsection{Knowledge distillation-based semi-supervised learning}
Semi-supervised learning has been widely explored for semantic and instance segmentation tasks that require costly pixel-level annotations \cite{yang2022survey}.
Among representative SSL strategies, pseudo labeling assigns model predictions to unlabeled images as supervision and has become a practical approach for segmentation tasks with limited annotated data.
However, pseudo labels often contain inaccurate boundaries, incomplete object regions, and missed instances due to the limited representation capability of the initial model.
Therefore, improving pseudo label quality remains a central issue in SSL for segmentation.

Knowledge distillation transfers rich prediction knowledge from a large teacher model to a compact student model and has been widely adopted for model compression \cite{hinton2015distilling}.
In SSL segmentation, teacher predictions can serve as pseudo labels for unlabeled images, allowing compact student models to learn from more diverse environments.
However, when the teacher is trained with limited labeled data, biased predictions can be propagated to the student through pseudo labels.
To address this limitation, recent studies \cite{yan2023sam4udass,yoon2025s4m} have adopted segmentation foundation models such as SAM \cite{kirillov2023segment} to refine pseudo masks with strong spatial precision.
These approaches have also been extended to teacher assistant and distillation strategies \cite{zhang2024goodsam,chen2025conformalsam} for compact student training and capacity gap reduction.
However, existing studies have generally treated foundation model based refinement and model compression as separate objectives, leaving their joint use in system level applications underexplored.
In this study, we propose a unified training pipeline that jointly performs SSL and model compression under limited CC domain annotations by distilling instance-level knowledge from an MTL teacher and foundation models.

\subsection{Multi-task learning for dense prediction}
Multi-task learning aims to learn shared representations across multiple dense prediction tasks within a unified model \cite{vandenhende2021multi}.
Dense prediction tasks, such as segmentation and monocular depth estimation, require rich pixel-level representations for semantic region understanding, instance separation, and geometric distance.
To obtain task-agnostic representations, previous MTL methods have relied on pixel-level annotations, high-capacity shared encoders, or cross-task interaction modules \cite{xu2018pad,vandenhende2020mti,ye2022inverted,ye2023taskprompter}.
This introduces two practical limitations, namely the annotation cost of supervised representation learning and the computational cost of deploying high-capacity MTL models.
These limitations impose a greater burden on real-world model deployment such as collision avoidance systems.

Recent MTL methods have attempted to reduce the computational cost of dense prediction by simplifying cross-task modules or applying distillation strategies.
Yang et al. \cite{yang2024multi} reduce the overhead of mixture-of-experts (MoE)-based MTL by incorporating low-rank experts and re-parameterized task heads.
Hwang et al. \cite{hwang2025meco} further improve efficiency by replacing explicit expert computation with nearest codebook retrieval through vector quantization.
KD-based methods have transferred task-agnostic knowledge from large teachers to smaller models, although their evaluation is still mainly limited to benchmark settings \cite{xu2023multi}.
In this study, we propose a practical training pipeline that combines SSL and KD to address both annotation cost and model complexity in the target real-world problem.

\begin{figure*}
	\centering
	\includegraphics[width=\textwidth]{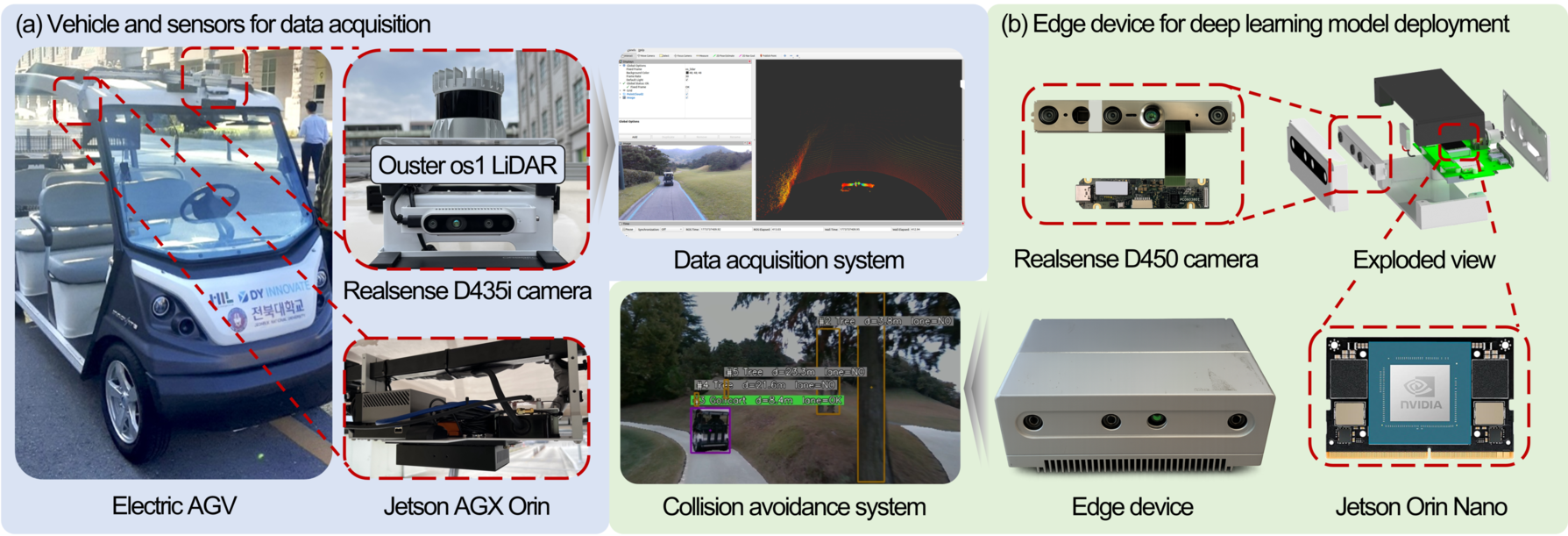}
	\caption{Sensor configurations of the AGV-based data acquisition platform and the edge device for on-board model deployment.}
	\label{fig1}
\end{figure*}

\section{Materials}
\label{Materials}

\subsection{Data acquisition platform}
We mounted multiple sensors on an electric AGV for driving data acquisition in CC environments, and the sensor configuration is shown in Fig. \ref{fig1} (a).
The maximum speed of the AGV is 20 km/h, and the sensors were mounted on a commercialized vehicle considering the camera position for deep learning model inference.
The sensor types include an Intel RealSense D435i camera and an Ouster OS1 LiDAR for image and point cloud acquisition, respectively.
In addition, GPS and IMU data were collected for trajectory tracking and attitude measurement, respectively, and all data were synchronized based on timestamps.
The robot operating system (ROS) middleware was used to organize the collected data, and the NVIDIA Jetson AGX Orin served as the computing platform.
Regarding the LiDAR specifications, the vertical field of view (FoV) ranges from $0^\circ$ to $-22.5^\circ$, which is smaller than the vertical FoV of the RGB camera (42.5$^\circ$) to satisfy the vertical coverage required for outdoor driving environment.

\subsection{Edge device}
We constructed an edge device to deploy the trained deep learning model and process inference results in the form of controller area network (CAN) messages, as illustrated by the exploded view in Fig. \ref{fig1} (b).
The edge device consists of an Intel RealSense D450 module for image acquisition, an NVIDIA Jetson Orin Nano for deep learning model inference, and a CAN software module for message transmission.
To minimize the cost of a production-grade AGV, the deep learning model and the overall system were designed for a low-cost embedded board.
Since the camera types used for data acquisition and the collision avoidance system are different, there is a discrepancy in the FoV of the RGB cameras.
Therefore, an additional image warping is required in preprocessing step.
As distance measurement is based on the estimated depth map from the deployed MTL model, additional sensors such as LiDAR, radar, or time of flight (ToF) cameras are not required.
Although the RealSense camera can obtain a dense depth map via a stereo setup, its depth reliability beyond 6 m is insufficient for collision avoidance systems in outdoor CC environments.
The visualization result of the collision avoidance system shows status of objects and semantic scene in real-time.

\begin{figure*}
	\centering
	\includegraphics[width=\textwidth]{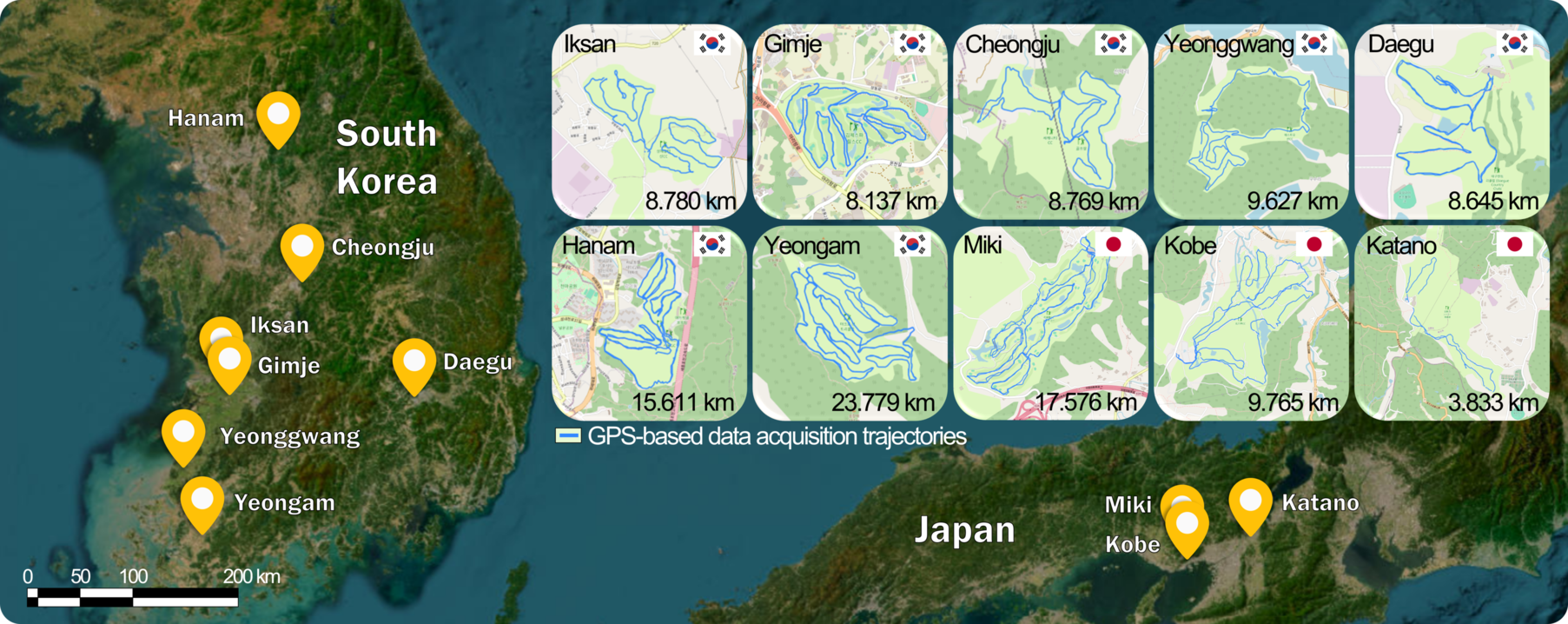}
	\caption{Visualization of the real-world data acquisition locations and driving trajectories across CCs in South Korea and Japan.}
	\label{fig2}
\end{figure*}

\subsection{Country club dataset}
We collected a real-world dataset to train the deep learning model and conducted driving experiments across ten CCs, including seven in South Korea and three in Japan as shown in Fig. \ref{fig2}.
The data collection period spans from September 4, 2023 to November 28, 2024.
Each season exhibits distinct characteristics, such as roads being occluded by fallen leaves in autumn and differences in vegetation in winter.
The weather conditions included five clear, three cloudy, and two rainy cases.
In clear conditions, excessive highlights are often observed, while rainy conditions introduce natural blur and occlusion caused by raindrop.
To incorporate natural reflection noise, the camera was mounted inside the windshield in six CCs, while an external mounting configuration was used for the others.
Driving sequence cases are defined for each CC, where teeing area serves as the primary split criterion.
The total driving distance is approximately 114.72 km across the CCs.

\begin{table}[t]
\centering
\caption{Overview of the CC dataset configuration at the video, image, and instance levels. The train and test sets were split at the level of independent sequences, and the road class was evaluated only in semantic segmentation.}
\label{tab1}
\resizebox{0.8\columnwidth}{!}{%
\renewcommand{\arraystretch}{1.4}
\begin{tabular}{cl|ccc}
\hline
Level & Category            & Total & Train & Test \\ \hline
\multirow{10}{*}{Video}    & Iksan      & 10    & 7     & 3    \\
                           & Gimje      & 13    & 7     & 6    \\
                           & Cheongju   & 9     & 7     & 2    \\
                           & Yeonggwang & 12    & 8     & 4    \\
                           & Daegu      & 26    & 24    & 2    \\
                           & Hanam      & 23    & 21    & 2    \\
                           & Yeongam    & 30    & 26    & 4    \\
                           & Miki       & 47    & 29    & 18   \\
                           & Kobe       & 39    & 39    & --   \\
                           & Katano     & 60    & 60    & --   \\ \hline
\multirow{3}{*}{Image} & Labeled depth ($D_d$)        & 45294 & 38352 & 6942 \\
                           & Unlabeled seg ($D_u$)       & 37478  & 37478   & --  \\
                           & Labeled seg ($D_l$)       & 1153  & 874   & 279  \\ \hline
\multirow{7}{*}{Instance} & Person     & 741    & 518   & 223     \\
                           & AGV        & 1412   & 1038  & 374     \\
                           & Tree       & 4047   & 2787  & 1260     \\
                           & Pond       & 176    & 144   & 32     \\
                           & Bunker     & 604    & 420   & 184     \\
                           & Unknown & 859   &  640  & 219     \\ \hline

\end{tabular}
}
\end{table}

\subsection{Data and annotation details}
\label{sec:annotataion}
The driving sensor data were synchronized and collected at a frequency of 10 Hz, and labels were generated to train each decoder of the dense prediction tasks.
The labels for depth estimation were generated from LiDAR point clouds, and camera–LiDAR calibration \cite{tsai2021optimising} was applied to obtain a sparse depth map at the same resolution as the image.
Camera–LiDAR calibration is performed based on the transformation matrix estimated from checkerboard images captured by both sensors, and this process is repeated whenever the sensor setup is changed.
Segmentation labels were generated using the computer vision annotation tool (CVAT) \cite{sekachev2019computer} in a two-stage process.
First, positive and negative points of object regions were provided as prompts to SAM to obtain coarse polygon masks.
Subsequently, improperly masked regions, merging errors, and boundary inconsistencies were manually refined in detail.
The primary instance classes in the CC environment include person, AGVs, hard part of trees, ponds, bunkers, and unknown objects (Unknown).
In addition, the semantic classes include road and background (BG) to cover the remaining regions of the driving scene.
Table \ref{tab1} provides an overview of the dataset, including video-level case splits, as well as the sizes of labeled and unlabeled datasets, and instance-level distributions.

\begin{figure*}
	\centering
	\includegraphics[width=\textwidth]{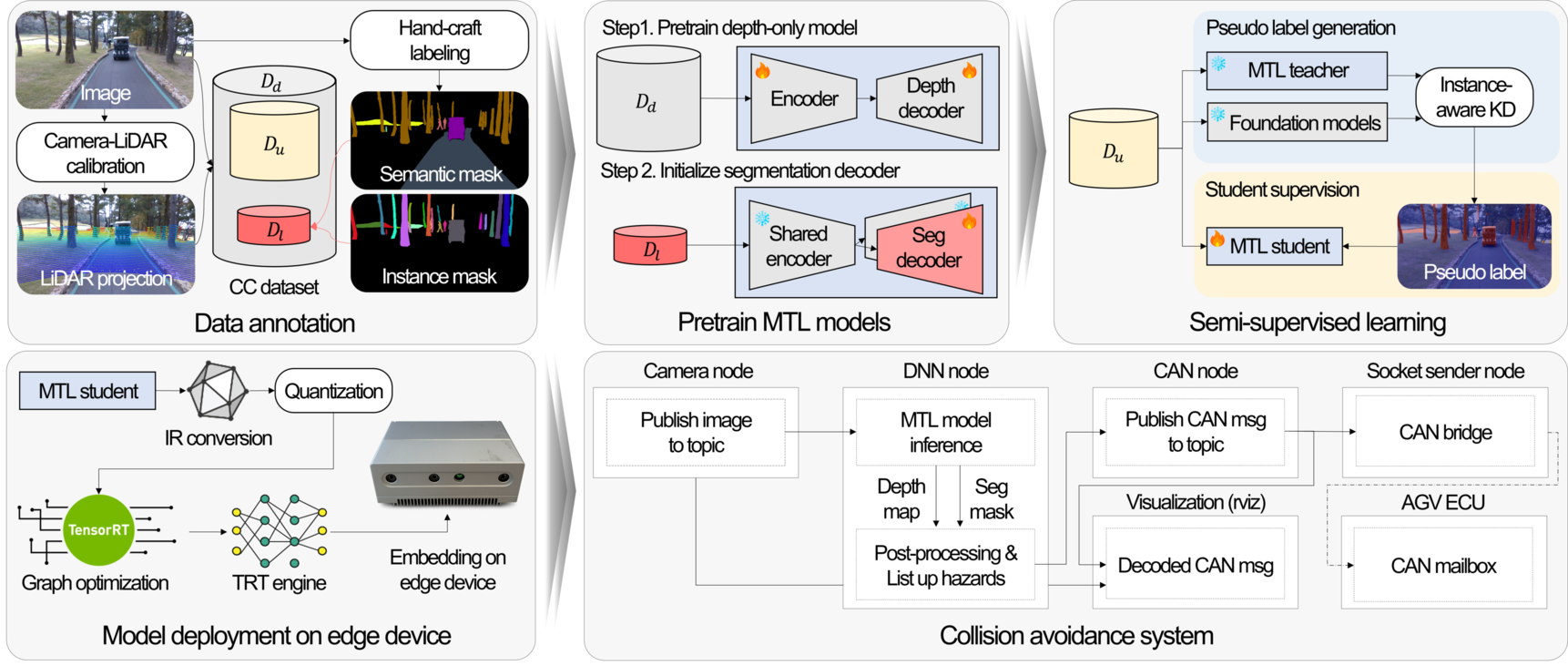}
	\caption{Overview of the proposed SSL training pipeline and collision avoidance system.}
	\label{fig3}
\end{figure*}

\section{Methodology}
\label{Methodology}
\subsection{Overall framework}
We propose a training framework for an MTL model and a real-time inference pipeline on an edge device for a collision avoidance system, as illustrated in Fig. \ref{fig3}.
The overall process consists of five stages: data annotation, pretraining of MTL models, semi-supervised learning, model deployment on the edge device, and operation of the collision avoidance system.
The datasets used for training the MTL model consist of $D_d$, $D_u$, and $D_l$, which denote the depth estimation dataset, unlabeled segmentation dataset, and labeled segmentation dataset, respectively, and the detailed data annotation stage is described in Section \ref{sec:annotataion}.
In the pretraining stage, the encoder and task decoders of the teacher and student MTL models are trained in two supervised learning steps on the labeled datasets $D_d$ and $D_l$ to obtain pretrained parameters.
In the SSL stage, pseudo labels are generated from unlabeled dataset $D_u$ through instance-aware knowledge distillation based on pixel-level probabilities derived from the instance-centric knowledge of the teacher and foundation models, and used as supervision for training the lightweight student.
The trained deployable student is further compressed and optimized for deployment on the edge device.
In the collision avoidance system, the deployed model converts pixel-level object information into CAN messages in real-time and transmits them to external controllers.
The detailed designs of instance-aware knowledge distillation, model deployment, and the collision avoidance system are described in Sections \ref{sec:instance-aware}, \ref{sec:model_deployment}, and \ref{sec:CAS}, respectively.

\subsection{Multi-task learning model architecture}
The MTL model adopted in this study is the mixture of low-rank experts (MLoRE) \cite{yang2024multi}, a MoE-based MTL model with a shared encoder architecture. 
MLoRE demonstrates strong performance on multiple dense prediction tasks by leveraging the powerful representation capability of vision transformer (ViT) \cite{dosovitskiy2020image} and routing expert-specific features to task decoders.
Moreover, the expert networks are designed based on low-rank adaptation \cite{hu2022lora} and the re-parameterization \cite{ding2021diverse} of parallel branches, leading to lightweight structure and improved inference speed compared to conventional MoE-based MTL models.
The original MLoRE addressed the dense prediction tasks such as depth estimation and semantic segmentation, which cannot provide instance-level discrimination within the same class region.
However, the proposed system targets instance-aware depth estimation and requires additional cues for object-level separation.
To this end, an instance decoder is designed by integrating an additional mask head into the semantic segmentation decoder.
The mask head outputs embeddings and is train based on discriminative loss, which minimizes the distance between instance clusters for each class.

\begin{figure*}
	\centering
	\includegraphics[width=\textwidth]{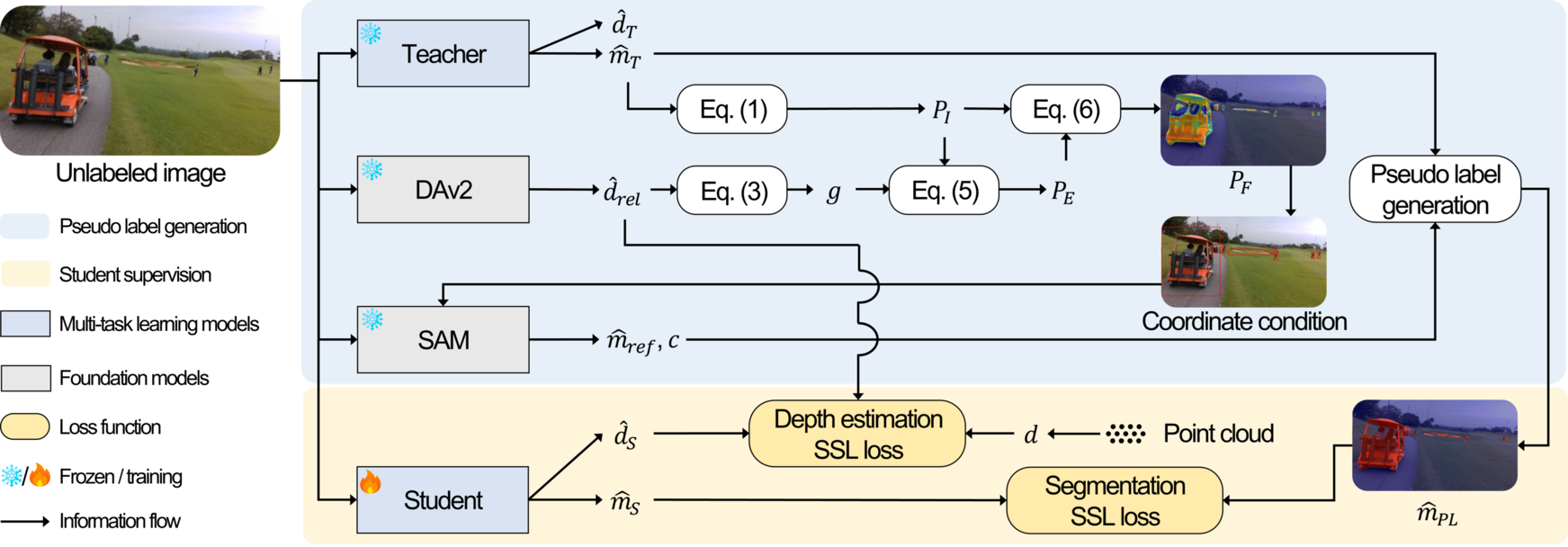}
	\caption{Overview of the proposed instance-aware KD pipeline for pseudo label generation and student supervision. Eq. denotes the equation numbers in this paper.}
	\label{fig4}
\end{figure*}

\subsection{Instance-aware knowledge distillation}
\label{sec:instance-aware}
To improve the generalization ability and achieve compression of the MTL model, we design a pseudo label generation method for the unlabeled segmentation dataset $D_u$ and a knowledge distillation pipeline, as illustrated in Fig. \ref{fig4}.
The proposed instance-aware KD adopts a multi-teacher framework that combines prior knowledge from a large MTL teacher trained on CC environments with foundation models that complement object characteristics.
The MTL teacher provides probabilities for instance-region of interest in the target scenario, while DAv2 provides edge probabilities for separating adjacent objects based on a precise relative depth map.
These two probabilities serve as the basis for deriving coordinate conditions for each object, which are then employed to obtain refined instance masks from SAM.
Finally, instance and semantic pseudo labels for student supervision are generated based on the intersection over union (IoU) between the teacher predictions and the SAM masks.

Specifically, the instance probability $P_I\in\mathbb{R}^{H \times W}$ provides coarse prior knowledge from the initialized teacher by separating foreground objects from semantic stuff.
$P_I$ is computed from the semantic prediction $\quad \hat{m}_T\in\mathbb{R}^{C \times H \times W}$ of the teacher and represents the pixel-level probability of instance regions, as defined in the following equation.
\begin{equation}
\tilde{P}_I = 1 - \sum_{k \in \mathcal{S}} Softmax \left(\frac{\hat{m}_T}{\tau}\right)_k, \quad \mathcal{S} = \{0, 1\},
\end{equation}
\begin{equation}
P_I = \tilde{P}_I \cdot \mathbb{1}(\tilde{P}_I > \theta),
\end{equation}
\noindent where $\mathcal{S}$, $\tau$, and $\theta$ denote the set of stuff class distributions (0:$Background$, 1:$Road$), the temperature, and the threshold parameter for defining probability components region, respectively.
$\tau$ controls the smoothness of $P_I$, where a lower value produces a sharper two-dimensional probability distribution with clearer object boundaries.
Although a low-uncertainty $P_I$ clearly separates stuff regions from the object regions, it is limited in distinguishing adjacent objects that share boundaries in the two-dimensional image plane.

To address this limitation, we introduce the edge probability $P_E\in\mathbb{R}^{H \times W}$ to estimate object boundaries within each component-level region of $P_I$.
$P_E$ is derived from the depth foundation model DAv2 to leverage the three-dimensional depth cues of the relative depth map $\hat{d}_{rel} \in \mathbb{R}^{H \times W}$.
$P_E$ is defined based on the gradient magnitude $g$ of the log-transformed $\hat{d}_{rel}$ and z-score normalization, and statistics are defined as follows.
\begin{equation}
g = \left\| \nabla \log \hat{d}_{rel} \right\|_2,
\end{equation}
\begin{equation}
\mu_k = Mean_{P_I^k}(g), \quad
\sigma_k = Std_{P_I^k}(g),
\end{equation}
where $P_I^k$ denotes the $k$-th connected probability component extracted from the thresholded $P_I$, representing a coarse instance region.
Subsequently, component-level z-score normalization is defined as follows.
\begin{equation}
P_E = \sum_{k \in \mathcal{B}} P_I^k \cdot Sigmoid \left( \frac{g - \mu_k}{\sigma_k} \right),
\end{equation}
\noindent where $\mathcal{B}$ denotes set of probability components, and $\mu_k$ and $\sigma_k$ represent the mean and standard deviation of $g$ within the $k$-th component region, respectively.
As a result, $P_E$ assigns higher scores to object boundaries when multiple connected probabilities of different objects are present in $P_I$.

SAM is a foundation model that takes an image and instance-aware prompts as input, and predicts pixel-level masks of the corresponding objects.
The teacher model is trained on a relatively small-scale CC dataset $D_l$, and has limited segmentation-aware representation.
In contrast, SAM is trained on diverse datasets and has strong generalization capacity for objects and scenes.
To generate coordinate conditions for SAM, we derive a final form of the probability map by integrating $P_I$ and $P_E$.
Since $P_I$ alone may produce merged components for adjacent objects, we fuse it with $P_E$ to suppress boundary regions and obtain spatially separated probability components.
The fused probability $P_F\in\mathbb{R}^{H \times W}$ is based on $P_I$ and $P_E$ extracted from the teacher and DAv2, which can be formulated as follows.
\begin{equation}
\tilde{P}_F = P_I \odot (1 - P_E)
\end{equation}
\begin{equation}
P_F = \tilde{P}_F \cdot \mathbb{1}(\tilde{P}_F > \theta),
\end{equation}
where $\odot$ denotes element-wise multiplication, and $P_F$ represents a probability map in which object regions of interest are separated in unlabeled images.
The coordinate condition is defined as the bounding boxes of the connected probability components extracted from $P_F$, and expanded by 10\% to mitigate underestimation by the teacher model.

\subsection{Loss functions}
We design SSL losses for instance segmentation and depth estimation based on instance-centric knowledge from pseudo label generation.
Since the pseudo labels generated from instance-aware KD depend on the search capability of the MTL teacher and the spatial mask precision of SAM, they may produce unreliable results.
In the proposed SSL pipeline, SAM evaluates the likelihood of objects within the prompt regions proposed by the MTL teacher and outputs confidence scores for the generated masks.
We utilize the average SAM confidence $c$ as a metric to assess the validity of the refined SAM mask $\hat{m}_{ref}$ and impose penalties on uncertain pseudo labels in the loss function, which can be defined as follows.
\begin{equation}
L_{sem} = c \, L_{CE}(\hat{m}_{PL}^{sem}, \hat{m}_{S}^{sem})
+ (1 - c) \, L_{KL}(\hat{m}_{T}, \hat{m}_{S}^{sem}),
\end{equation}
\begin{equation}
L_{inst} = c \, L_{disc}(\hat{m}_{PL}^{inst}, \hat{m}_{S}^{inst}),
\end{equation}
where $sem$ and $inst$ denote semantic and class-agnostic instance segmentation, respectively, while $L_{CE}$, $L_{KL}$, and $L_{disc}$ represent cross-entropy, Kullback–Leibler divergence, and discriminative loss.
In addition, subscripts $T$, $S$ and $PL$ denote the teacher prediction, student prediction, and pseudo label, respectively.
$\hat{m}_{PL}^{sem}$ is generated from $\hat{m}_{ref}$ via IoU matching with $\hat{m}_{T}$, while $\hat{m}_{PL}^{inst}$ is constructed from its confidence-ranked masks.

For depth estimation, the sparsity of the LiDAR projection $d$ and the limited FoV coverage create regions that cannot be supervised during training, which degrades overall consistency of depth map.
On the other hand, the relative depth map $\hat{d}_{rel}$ provides dense and global depth information, although its scale does not necessarily correspond to accurate metric depth.
Based on this property, we further utilize the trend information of $\hat{d}_{rel}$ to improve reliability in outside regions of the FoV coverage, and the overall SSL loss function for depth estimation is defined as follows.
\begin{equation}
L_{dep} = L_{silog}(d, \hat{d}_S) + \omega \left( 1 - \rho(\hat{d}_{rel}, \hat{d}_S) \right),
\end{equation}
where $L_{dep}$, $L_{silog}$, $\rho(\cdot)$, and $\omega$ denote depth estimation, scale-invariant log, Pearson correlation, and the loss weight, respectively.
The final SSL loss function for the lightweight MTL model is defined as $L_{total} = \sum_{t \in \mathcal{T}} \lambda_t L_t$, where $\mathcal{T}$ and $\lambda$ denote the set of target tasks $t$ and the task balancing weight, respectively.
Each task loss $L_t$ is computed based on LiDAR projection $d$ and pseudo labels generated by instance-aware KD, and $L_{total}$ integrates multiple supervision sources to jointly optimize all target tasks.

\subsection{Model deployment}
\label{sec:model_deployment}
After semi-supervised learning, the MTL student is deployed on an edge device for real-time inference in the AGV platform.
To mitigate the computational constraints of the edge device, the student is transformed into a deployable form through intermediate representation (IR) conversion and graph-level optimization.
In the graph-level optimization stage, additional compression processes such as layer fusion and half-precision quantization are applied to reduce memory usage and computational cost.
Since the edge device is built on an NVIDIA embedded platform, the developed deep learning model is converted into a TensorRT (TRT) engine format.
The generated TRT engine is deployed in the DNN node of the collision avoidance system based on ROS2 \cite{macenski2022robot} middleware, and performs real-time inference of dense prediction results, including depth maps and segmentation masks.

\begin{table}[t]
\centering
\caption{Byte-level CAN payload definition of the object descriptor.}
\label{tab2}
\resizebox{\columnwidth}{!}{%
\renewcommand{\arraystretch}{1.4}
\begin{tabular}{c|l|c|l}
\hline
Byte idx & Field & Enc. & Description \\ \hline
1 & Obj\_ID      & 1--$K$   & Distance-ordered object ID \\
2 & Obj\_cls     & 2--8   & Object class \\
3 & Lane\_status & 1/2    & Road-area adjacency flag \\
4 & Dist. (cm)   & uint8  & Normalized distance \\
5 & $x$-center     & uint8  & Normalized bbox center $x$ \\
6 & $y$-center     & uint8  & Normalized bbox center $y$ \\
7 & Width        & uint8  & Normalized bbox width \\
8 & Height       & uint8  & Normalized bbox height \\ \hline
\end{tabular}
}
\end{table}  

\subsection{Collision avoidance system}
\label{sec:CAS}
The collision avoidance system is shown as the final process in Fig. \ref{fig3}, where image acquisition, model inference, post processing, and CAN message transmission are organized in a ROS2 pipeline.
The camera node publishes input images to the image topic, and the DNN node performs MTL model inference to obtain depth maps and segmentation masks.
The dense prediction results of the deployed model are post processed into object descriptors and converted into a CAN protocol format for transmission to external electronic control units (ECU).
As the AGV operates in a CAN 2.0 environment, each object is encoded as an object descriptor composed of eight byte-level fields, and the assigned values are presented in Table \ref{tab2}.
The object region is represented in bounding box form obtained from the minimum bounding rectangle enclosing the instance mask, and the object category is determined based on the class that achieves the maximum IoU with the semantic mask.
The distance to each object is computed by averaging the minimum points corresponding to the lower quantile of the depth distribution within the instance mask.
The Lane\_status field serves as a flag for object adjacency and is activated when the corridor region below the bounding box overlaps with the road area.
The object descriptors are sorted in ascending order of the distance to ensure that the ECU can prioritize receiving and processing nearby hazardous objects under the limited bus payload constraints of CAN 2.0.
The coordinate, size, and distance values are normalized to a range of 0–255 based on their respective maximum values and encoded as 8-bit integers.

The CAN node is designed to convert each object descriptor into CAN frame format and publish it onto the CAN bus. 
To this end, each CAN frame is sequentially transmitted in a round-robin manner within a fixed transmission cycle to ensure stable transmission of multiple object information.
The socket sender node relays CAN frames from ROS2 to the ECU communication channel as a bridge.
The visualization module operates by decoding the transmitted CAN messages and displaying the object information to observe false positives and missed detections in the final perception results.
Finally, the AGV ECU receives CAN messages from the edge device and interprets information on nearby objects and drivable areas to perform warning and control actions.

\section{Experimental results}
\label{Experimental results}
\subsection{Experimental setup}
The entire experiments for training the MTL student were conducted on a hardware system equipped with Intel Core i9-10940X CPU, 64 GB DDR4 RAM, and dual NVIDIA GeForce RTX 3090 Ti GPUs.
The trained student was deployed on NVIDIA Jetson Orin Nano-based edge device for inference and field validation.
The experiments were conducted under the same software environment on both the edge device and the server, including Python 3.8.18, PyTorch 1.11, and TensorRT 8.5.2.
In particular, the edge device operated under Ubuntu 20.04, JetPack 5.1.2, CUDA 11.4, and cuDNN 8.6.0.
The data acquisition and the collision avoidance system utilized ROS and ROS2 middleware, respectively.

The MTL teacher and student employed ViT-Large and ViT-Tiny backbones to learn global task-agnostic representations.
Additionally, ViT-Huge and ViT-Large backbones were applied to SAM and DAv2, to fully leverage the generalization capability of foundation models.
In the server environment, all models were trained with the Adam optimizer, with a batch size of 4, a learning rate of $1 \times 10^{-4}$, and a weight decay of $1 \times 10^{-6}$.
In the edge device environment for model inference, the input image resolution and batch size were set to $640 \times 352$ and 1, considering computational constraints.
All performance evaluations were conducted on the same labeled test dataset under identical execution conditions to ensure fair comparison.
Due to the limited size of the labeled test dataset, all models were evaluated at fixed 70 epochs.

\subsection{Evaluation metrics}
We evaluated the performance of the proposed method from the perspectives of semantic segmentation, instance segmentation, and depth estimation tasks.
The segmentation decoder of the MTL model is designed to jointly estimate a semantic mask and a class-agnostic instance mask to perform instance and stuff-class segmentation simultaneously.
Accordingly, segmentation performance is evaluated separately by distinguishing the roles of the two prediction masks.
Specifically, semantic segmentation performance is measured by mean and class-wise IoU to evaluate pixel-level prediction accuracy across all classes in the driving scene.
In contrast, class-agnostic instance segmentation performance excludes the stuff-class set and is measured with Precision, Recall, and F1-score to evaluate separation and detection of individual object regions.
These metrics jointly reflect detection accuracy and failures, making them suitable for evaluating individual object perception performance, which is critical for collision avoidance systems.
Depth estimation performance is evaluated on valid points using absolute relative error (Abs Rel), squared relative error (Sq Rel), root mean squared error (RMSE), and root mean squared logarithmic error (RMSE $log$) to analyze the discrepancy between predicted and GT depth maps.

\subsection{Quantitative segmentation results}
We evaluated the segmentation performance of the deployable student under various baseline methods on a real-world CC dataset, and Tables \ref{tab3} and \ref{tab4} present the results for semantic and class-agnostic instance segmentation, respectively.
In the semantic segmentation results, the teacher achieved an mIoU of 72.47\%, whereas the plain student showed a significant performance gap with an mIoU of 37.72\%.
However, applying baseline KD guided by $\hat{m}_T$, the mIoU was significantly improved to 70.23\%, which was close to the teacher performance.
Moreover, incorporating SSL with unlabeled data $D_u$, improved the mIoU to 70.47\%, slightly surpassing the baseline KD.
With the proposed SSL method, the mIoU further increased to 75.65\%, achieving the best performance and surpassing the teacher performance.
This trend was also consistent in the class-wise results, with notable improvements observed in classes such as person, tree, pond, bunker, and unknown objects.
However, the performance on pond and unknown object classes remained relatively low, due to data scarcity and label inconsistency.

Similarly, in the class-agnostic instance segmentation, the teacher achieved an F1-score of 39.32, whereas the plain student exhibited substantially lower performance of 18.22.
The baseline KD and SSL settings significantly improved the F1-score to 42.11 and 44.28, respectively, both exceeding that of the teacher.
The student trained with the proposed method outperformed all baseline methods, achieving an F1-score of 51.46, a Precision of 59.34, and a Recall of 45.43.
Specifically, the substantial improvement in Recall from 35.71 to 45.43 indicates a reduction in false negatives for existing objects achieved by the proposed SSL method.
These results suggest that instance-centric knowledge from the foundation model can refine imprecise teacher predictions into spatially precise pseudo labels.
The proposed method can improve both semantic scene understanding and instance-level object separation.

\begin{table*}[t]
\centering
\caption{Quantitative results on semantic segmentation.}
\renewcommand{\arraystretch}{1.4}
\resizebox{0.75\textwidth}{!}{%
\begin{tabular}{cccc|cccccccc|c}
\hline
$D_l$ & $D_u$ & $\hat{m}_T$ & $\hat{m}_{ref}$ & BG & Road & Person & AGV & Tree & Pond & Bunker & Unknown & mIoU \\ 
\hline
\multicolumn{4}{c|}{Teacher} 
& 95.93 & 94.04 & 62.85 & 90.38 & 60.61 & 60.61 & 63.57 & 51.83 & 72.47 \\ 
\hline
\checkmark &          &          &           & 93.44 & 89.85 & 0.39  & 79.05 & 38.97 & 0.00 & 0.07 & 0.00 & 37.72 \\
\checkmark &          & \checkmark &           & 96.09 & 94.69 & 60.23 & 90.48 & 59.51 & 51.63 & 60.22 & 49.06 & 70.23 \\
\checkmark & \checkmark & \checkmark &           & 96.36 & 95.13 & 62.39 & 90.83 & 61.09 & 45.50 & 68.15 & 44.33 & 70.47 \\
\checkmark & \checkmark & \checkmark & \checkmark 
& 96.77 & 95.55 & 65.08 & 91.96 & 67.05 & 56.43 & 76.96 & 55.43 & 75.65 \\
\hline
\end{tabular}%
}
\label{tab3}
\end{table*} 

\begin{table}[t]
\centering
\caption{Quantitative results on class-agnostic instance segmentation.}
\label{tab4}
\resizebox{0.8\columnwidth}{!}{%
\renewcommand{\arraystretch}{1.4}
\begin{tabular}{cccc|ccc}
\hline
$D_l$ & $D_u$ & $\hat{m}_T$ & $\hat{m}_{ref}$ & Precision & Recall & F1 \\ \hline
\multicolumn{4}{c|}{Teacher} & 50.89 & 32.03 & 39.32 \\ \hline
\checkmark &          &          &           & 30.40 & 13.01 & 18.22 \\
\checkmark &          & \checkmark &           & 53.95 & 34.54 & 42.11 \\
\checkmark & \checkmark & \checkmark &           & 58.29 & 35.71 & 44.28 \\
\checkmark & \checkmark & \checkmark & \checkmark  & 59.34 & 45.43 & 51.46\\ \hline
\end{tabular}
}
\end{table} 

\begin{table}[t]
\centering
\caption{Quantitative results on monocular depth estimation.}
\label{tab5}
\resizebox{\columnwidth}{!}{%
\renewcommand{\arraystretch}{1.4}
\begin{tabular}{cccc|cccc}
\hline
$D_l$ & $D_u$ & $\hat{d}_T$ & $\hat{d}_{rel}$ & Abs Rel & RMSE & RMSE $log$ & Sq Rel \\ \hline
\multicolumn{4}{c|}{Teacher} & 0.0966 & 2.4511 & 0.1852 & 0.4184 \\ \hline
\checkmark &          &          &          & 0.1342 & 3.0183 & 0.2262 & 0.5776 \\
\checkmark &          & \checkmark &          & 0.1414 & 3.0718 & 0.2329 & 0.5952 \\
\checkmark & \checkmark & \checkmark &          & 0.1422 & 3.1777 & 0.2392 & 0.6121 \\
\checkmark & \checkmark &          & \checkmark & 0.1258 & 2.8941 & 0.2178 & 0.5361 \\ \hline
\end{tabular}%
}
\end{table} 

\subsection{Quantitative depth estimation results}
Table \ref{tab5} presents quantitative results of depth estimation for the MTL student models.
Teacher and all students were pretrained on the depth dataset $D_d$ and then fine-tuned with additional supervision sources such as the MTL teacher and DAv2.
The teacher achieved 0.0966 in Abs Rel, 2.4511 in RMSE, 0.1852 in RMSE $log$, and 0.4184 in Sq Rel, whereas the plain student showed substantial performance degradation with values of 0.1342, 3.0183, 0.2262, and 0.5776, respectively.
However, when teacher-based KD and SSL were applied, the overall error increased as the influence of teacher supervision became greater.
This suggests that imprecise teacher depth maps negatively affect the student, and if the unlabeled dataset size increases, the tendency grows more pronounced. 
In contrast, whereas the $\hat{d}_{rel}$ was used as an auxiliary supervision source, improvements were observed across all metrics with Abs Rel of 0.1258, RMSE of 2.8941, RMSE log of 0.2178, and Sq Rel of 0.5361.
These results indicate that learning the global trend of the precise relative depth map with Pearson correlation improves the overall coherence of depth map.

\subsection{Analysis of probabilities and pseudo labels}
\begin{figure*}
	\centering
	\includegraphics[width=\textwidth]{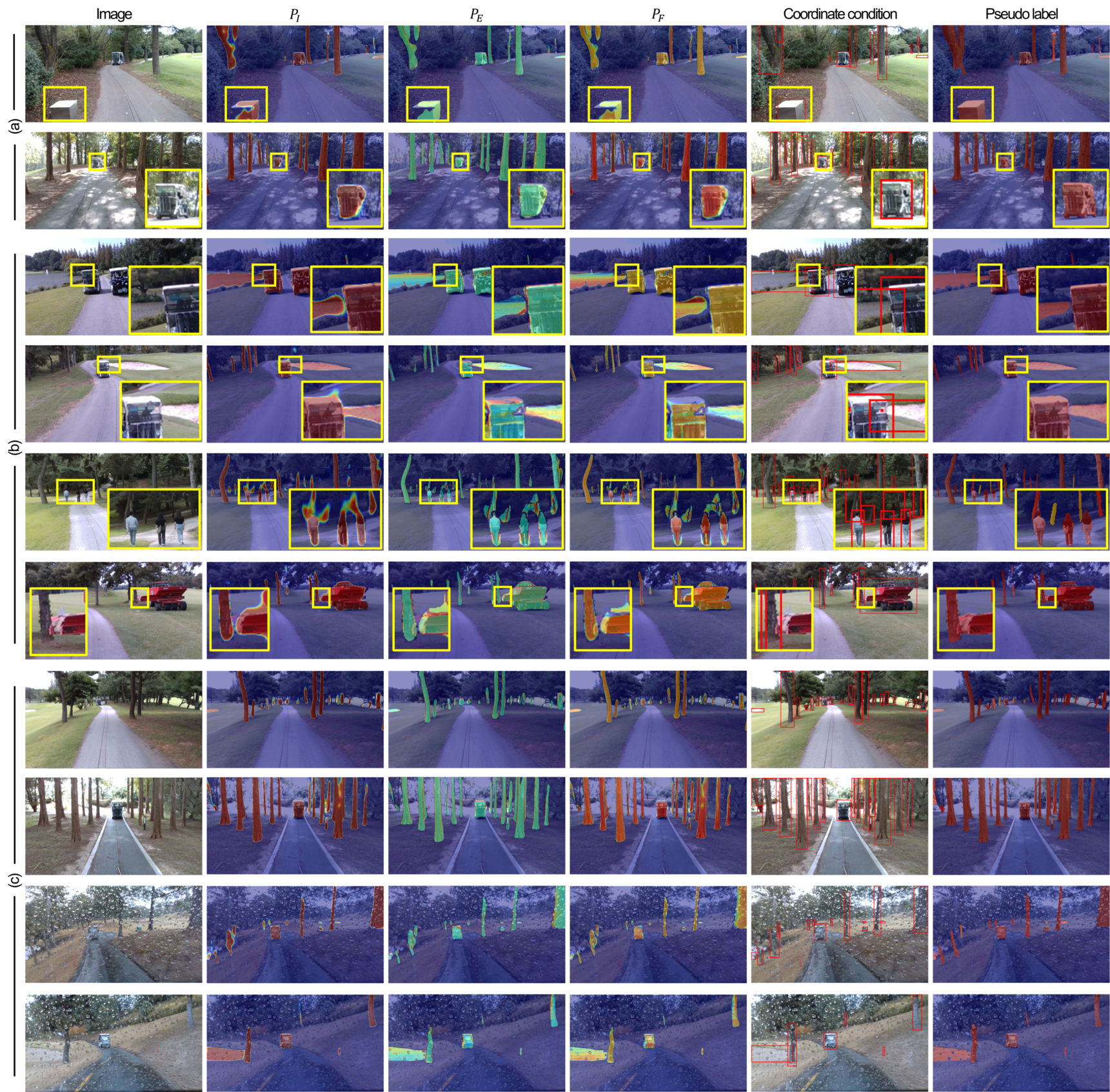}
	\caption{Visualization of probabilities, coordinate condition, and pseudo label, where colors from red to blue indicate high to low probability or confidence, respectively.}
	\label{fig5}
\end{figure*}
To explore the instance-aware KD process, we visualized the probabilities and pseudo label mask confidence, as shown in Fig. \ref{fig5}. 
The visualization results consist of front-view cases along with the input image, $P_I$, $P_E$, $P_F$, coordinate condition, and pseudo label masks, where the primary regions are highlighted in yellow boxes.
Case (a) illustrates the refinement of incomplete predictions after applying the proposed method, which can be observed through a comparison between $P_I$ and the pseudo label.
Specifically, in the first and second samples, spatially incomplete predictions were obtained for a nearby unknown object and an AGV on the road, respectively.
By employing them as conditions for the foundation model, high confidence pseudo labels were generated.
On the other hand, when conditions are generated solely from $P_I$, separation between adjacent components is compromised.
Consequently, SAM tends to infer only a single dominant object within the bounding box.
Case (b) demonstrates the effect of $P_E$, which suppresses the influence of high edge probability regions, allowing components of adjacent instances to be separated in $P_F$.
Finally, case (c) presents pseudo label generation results in challenging environments, such as multiple small objects, dense forests, or heavy rainfall.
Although instance-aware KD performed well in most challenging scenarios, for mispredictions such as the bunker region in the last sample, compensation with a teacher-based KD loss is necessary.

\begin{figure*}
	\centering
	\includegraphics[width=\textwidth]{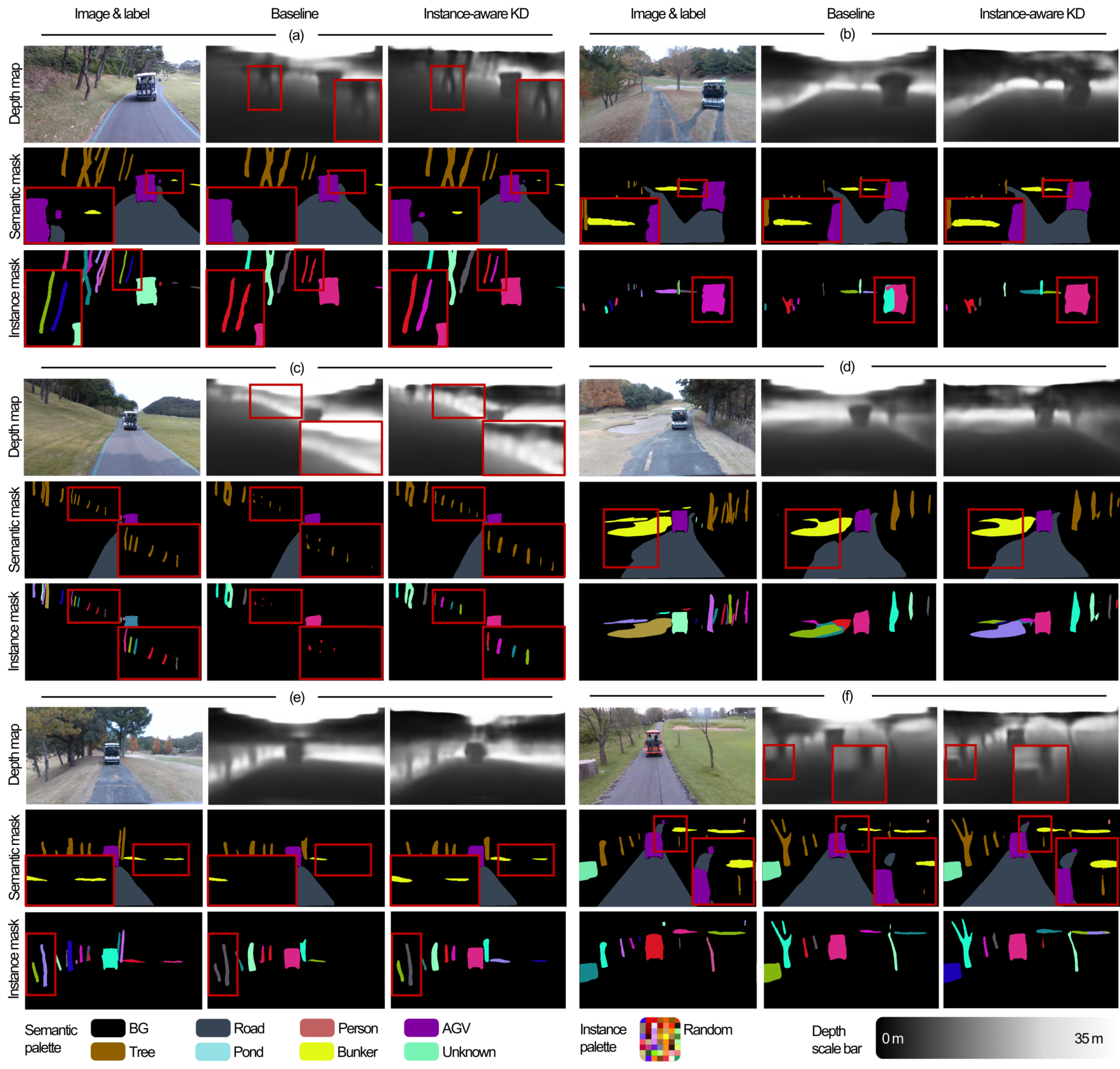}
	\caption{Comparison of qualitative task results on baseline and proposed SSL methods.}
	\label{fig6}
\end{figure*}

\begin{figure*}
	\centering
	\includegraphics[width=\textwidth]{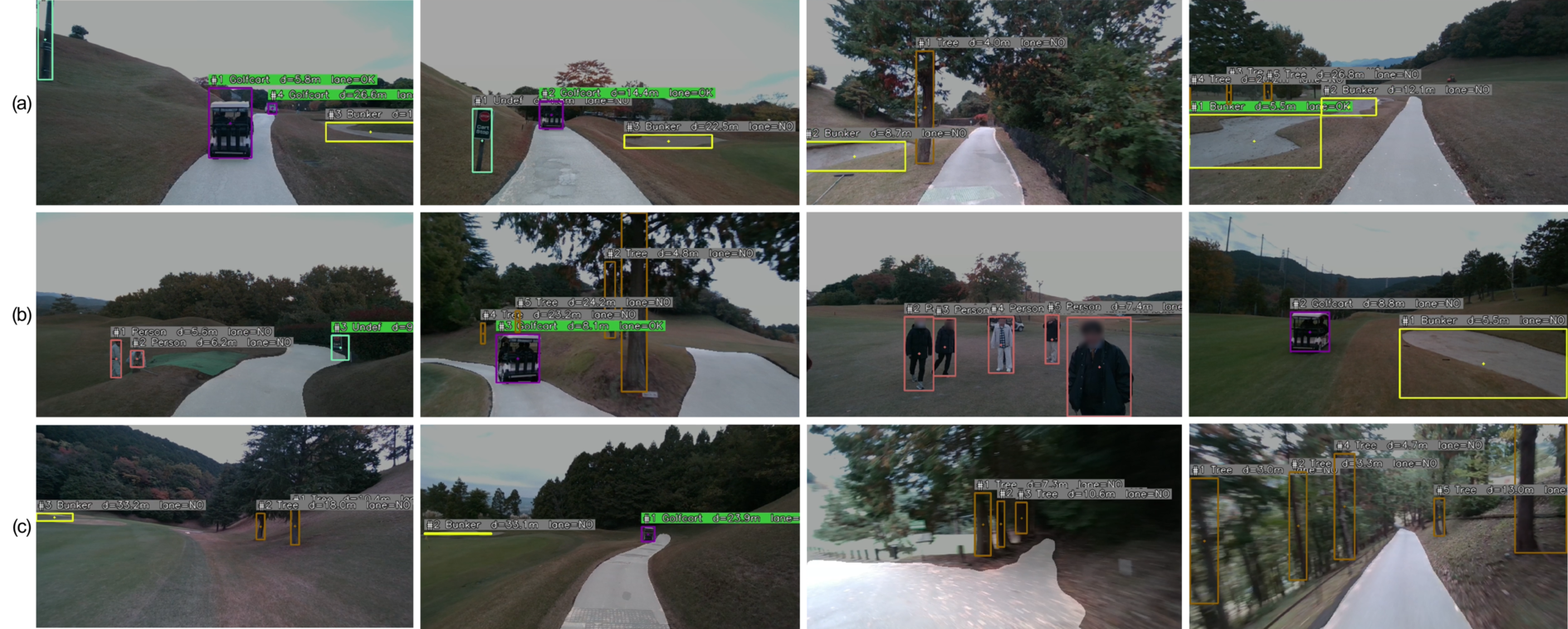}
	\caption{Visualization of decoded CAN messages and road masks in the proposed collision avoidance system.}
	\label{fig7}
\end{figure*}

\subsection{Qualitative results}
We compared the qualitative results of MTL student predictions obtained from the baseline SSL and the proposed method, as shown in Fig. \ref{fig6}.
The qualitative comparison consists of input images, labels, and dense predictions for each task across samples (a) to (f), with highlighted regions marked by red boxes.
For depth estimation, notable improvements in precision are observed in samples (a), (c), and (f), indicating the effectiveness of supervision based on the spatial depth trend from $\hat{d}_{rel}$.
In semantic segmentation, the effect of refined pseudo labels is prominent in samples (b) and (d), where spatially incomplete bunker regions are improved.
Moreover, in samples (a), (c), (e), and (f), the student of the instance-aware KD successfully captures small objects that are missed by the baseline method.
Additionally, in multiple samples such as (a), (b), (c), and (e), the results support the quantitative improvements observed in class-agnostic instance segmentation.

\subsection{Collision avoidance system results}
The CAN messages generated by the lightweight MTL model were analyzed during AGV operation in a real CC environment to validate the proposed collision avoidance system.
Fig. \ref{fig7} presents the input image and decoded CAN messages with the road area, considering only the five nearest objects.
Each object class is represented by denormalized bounding boxes with the same color scheme as the semantic segmentation palette in Fig. \ref{fig6}, while the road area is visualized at the pixel level in white.
The tag of each detected object includes the ID, class label, distance, and lane status flag, and is shown in green or gray when the flag is True or False, respectively.
Cases (a) and (b) illustrate scenarios with objects located near the road and the ego AGV, respectively, while case (c) shows challenging scenarios where objects are distant and small or affected by motion blur due to high-speed driving.
These results demonstrate that the proposed system enables immediate perception under diverse object distributions and challenging visual conditions during real-world CC operation. 

\subsection{Computational complexity}
Table \ref{tab6} presents a comparison of computational complexity and latency between the MTL teacher and student.
The teacher requires 1.28 T floating-point operations (FLOPs) and 1.29 B parameters, whereas the student shows a significantly lighter architecture with 56.55 G FLOPs and 90.40 M parameters.
As a result, the student reduces FLOPs by approximately 22.68$\times$ and the number of parameters by 14.33$\times$ compared to the teacher, achieving efficiency suitable for deployment on edge device.
In contrast, the teacher model is limited as a deployable solution due to its high computational cost and large model size.
In terms of latency, the student achieved 17.23 frames per second (FPS) in a server environment, which is approximately 2.03$\times$ faster than the teacher’s latency, 8.48 FPS.
Furthermore, the deployed student operated at 6.46 FPS for model inference and 5.17 FPS at the system level on the edge device, where the system includes preprocessing, postprocessing, and CAN transmission as well.
Here, the edge FPS was measured after graph-level optimization, including layer fusion and half-precision quantization.
These results demonstrate that the proposed method offers strong predictive performance while maintaining the computational efficiency required for practical edge deployment scenario.

\begin{table}[t]
\centering
\caption{Comparison of computational complexity and latency.}
\label{tab6}
\renewcommand{\arraystretch}{1.4}
\resizebox{0.9\columnwidth}{!}{%
\begin{tabular}{l|cccc}
\hline
Model & FLOPs & \#Params & Server FPS & Edge FPS \\ \hline
Teacher & 1.28 T & 1.29 B & 8.48 & -- \\
Student & 56.55 G & 90.40 M & 17.23 & 6.46 \\ \hline
\end{tabular}%
}
\end{table}

\section{Conclusion}
\label{Conclusion}
In this paper, we proposed a monocular camera-based multi-task dense prediction model and an instance-aware KD-based SSL framework for on-device collision avoidance system in CC driving environments.
The proposed method is designed to overcome limitations in computation and labeling cost by distilling instance-centric knowledge from multiple teacher models.
It combines instance priors from a large-scale teacher and precise general cues from foundation models to generate pseudo labels.
Experimental results show that the proposed method outperforms the teacher and baseline SSL method in segmentation, and demonstrates the effectiveness of relative depth priors in depth estimation.
The student achieves significantly reduced computational complexity and parameters compared to the teacher, while operating in real-time on low-cost edge device.
As a result, the proposed system reliably transmits obstacle location and distance information via CAN messages within an actual AGV control system.
This study integrates a real-world CC dataset, a lightweight MTL model, and an instance-aware KD-based SSL framework, thereby presenting a practical vision-based collision avoidance system.
Future work will focus on enhancing model compression techniques to improve system inference speed while maintaining multiple dense prediction functionality.

\printcredits

\section*{Declaration of competing interest}
The authors declare no competing financial interests or personal relationships that could have influenced the work reported in this paper.

\section*{Data availability}
The data are not publicly available due to confidentiality restrictions related to field data collected at country club sites. Data may be available from the corresponding author upon reasonable request and with permission from the participating organizations.

\section*{Funding}
This work was supported by the Institute of Information \& Communications Technology Planning \& Evaluation(IITP)-Innovative Human Resource Development for Local Intellectualization program grant funded by the Korea govern-\\ment(MSIT)(IITP-2026-RS-2024-00439292).

\bibliographystyle{elsarticle-num-names}

\bibliography{cas-refs}


\end{document}